# A multi-sourced data and agent-based approach for complementing Time Use Surveys in the context of residential human activity and load curve simulation

Mathieu Schumann[1], Quentin Reynaud[2], François Sempé[3], Julien Guibourdenche[4], Jean-Baptiste Ly[1,5], Nicolas Sabouret[5]
[1]EDF Lab Paris Saclay, Palaiseau, France
[2]QRCI, Clermont-Ferrand, France
[3]FSCI, Paris, France
[4]AKTEN, Fontaines-sur Saône, France
[5]LISN, Université Paris-Saclay, CNRS, Orsay, France

## Abstract

To address the major issues associated with using Time-Use Survey (TUS) for simulating residential load curves, we present the SMACH approach, which combines qualitative and quantitative data with agent-based simulation. Our model consists of autonomous agents assigned with daily tasks. The agents try to accomplish their assigned tasks to the best of their abilities. Quantitative data are used to generate tasks assignments. Qualitative studies allow us to define how agents select, based on plausible cognitive principles, the tasks to accomplish depending on the context. Our results show a better representation of weekdays and weekends, a more flexible association of tasks with appliances, and an improved simulation of load curves compared to real data.

## Highlights

- Discussion about Time-Use Surveys (TUS) limits and the use of TUS in activity and energy simulation
- Presentation of complementary data both qualitative and quantitative used to complement TUS data
- Proposition of an agent-based approach that balances these limitations

## Introduction

The world of energy has been undergoing structural changes notably associated with the electrification of energy uses and the insertion of local renewable energies into smart grids. The increased understanding of the challenges of climate change as well as the ongoing energy crises are also driving the need to accelerate the energy transition in the building sector. Indeed, buildings still account for 30% of the global final energy consumption in 2021, according to the IEA. However, this can only be achieved with the involvement of citizens at both the individual and collective scales. Performing human-centric assessments of energy consumption at the household and urban levels (Dabirian et al., 2022) is critical to make these changes operational.

The now-documented impact of human activity on energy consumption in buildings brought the research community together to tackle the various scientific issues arising from this topic, particularly within IEA EBC Annex 79 (Dong et al., 2021). An increasing number of approaches use Time Use Surveys (TUS) and stochastic individual-based approaches to generate occupancy profiles, model interactions with building systems, or evaluate energy consumption related to human activity. However, the state-of-the-art analyses Osman and Ouf (2021) identify limitations to using TUS to simulate coherent human activity and the associated energy consumption (Yamaguchi et al., 2019).

In the first section, we review and discuss these limitations. The second section presents our proposition to complement TUS data to effectively address these limitations. We show how qualitative studies on human activity in social sciences and cognitive ergonomics are essential to complement TUS data that are purely quantitative. Then we describe our human activity model based on Agent-Based Modelling (ABM) and how this paradigm allows us to tackle the discussed TUS limitations. We describe how this led to the improvement of the appliances use and energy consumption model and illustrate its validation process, with a focus on specific electricity. Finally, we demonstrate the practical applications of our model through two Use Cases.

## State of the art: limitations of TUS data and their use

McKenna et al. (2018) as well as Osman and Ouf (2021) list major issues regarding TUS data and their use. We summarize and discuss them in the following list:

### Limitation 1 (L1): no link between activities and energy use

Classical state-of-the-art approaches tend to oversimplify the relation between appliances and activities (e.g., by directly linking appliances to activities), leading to inaccurate and too static temporal appliance use profiles. This relation is highly individual-dependent and totally absent in TUS data. For example, the "laundry" activity in TUS data could be done with or without a washing machine; and the relation between the "cooking" activity and the concrete use of an oven is complex and impossible to model solely based on TUS data. TUS also fail to capture any potential energy consuming situations that are not directly linked to the current activity ("watching TV while eating" or "having a radio on in another room").

### L2: lack of data about variability and interactions between activities

TUS data lack explanation about sequences of activities and days because it tends to be collected for single days only. More generally, TUS data fail to capture the temporal activities variability: routine activities, temporal variations around routine activities, and exceptional activities. Classical state-of-the-art approaches tend to use these data to represent average days, neglecting the behavioral change due to external factors (weather, communal events, national holidays, etc.) as well as internal factors (mental or physical states, emotions, etc.). They usually fail to produce a consistent series of activities because they assume that activities are random variables that depend on a fixed number of independent variables (e.g., day of the week, total number of residents in a household, etc.) that do not vary in time. McKenna et al. (2018) point out that activities are interdependent and that time-shifting one can have complex impacts on others. They stress that understanding interdependencies and evaluating the impact of changing activity timings is crucial for demand response studies. The absence of variability in activity patterns is a significant concern for modeling human activity, as it is a crucial feature according to Attaianese and Duca (2012).

**L3: lack of data about the impact of individual characteristics of occupants, households, dwellings and appliances**

TUS-based approaches tend to average individuals and neglect factors that may have strong associations with energy demand (employment status, work patterns or whether children, elderly relatives, guests). The characteristics of an average resident may have little meaning for demand response purposes if few people, in practice, conform to the characteristics of the average. Similarly, appliances are often assumed to be average both in terms of technical characteristics and use which decreases the diversity of energy consumptions. This results in the simulation of households that are overly representative of the average, whether in terms of household members or their dwellings.

**L4: absence of collective activities or household-level organisation**

TUS data are focused on the individual level and makes it hard to detect the overlapping between household members while performing shared activities. Moreover, TUS data assume that residents perform one activity at a time and that activities are not combined (e.g., a continuous combination of childcare and watching TV is not a recognized state). However, activities and consumptions are the product of individual, collective and interdependent practices (Shove & Walker, 2014) resulting from complex household dynamics, but this higher level of organization is often overlooked in state-of-the art approaches (Yamaguchi and Shimoda, 2017).

**L5: biases in TUS**

Although TUS are using a proven methodology, they exhibit some major biases. Firstly, all short activities (<10 minutes) may have trouble of being captured in TUS data since TUS standards rely on 10-min episodes. "Boiling water with a kettle" or "opening a window" may be totally absent from TUS datasets. Moreover, Osman and Ouf (2021) underline the fact that TUS surveys are affected by the respondents' memory, accuracy, and willingness to round up or down the actual time spent on different activities, as well as by the social desirability and bias that might overestimate the time spent on some activities rather than other. In addition, these purely quantitative data tend to hide the fact that a single activity does not mean the same thing for different people (Poizat et al., 2009). "Gardening" could be a job or a leisure activity; "cooking" could be a solitary, time-pressed activity, considered a chore, or a collective and relaxing activity. Furthermore, TUS surveys are generally conducted every 10 years (but often less than that), which raises the question of the impact of potentially outdated data.

**State-of-the-art propositions to tackle these limits**

The literature reports numerous works that propose advancements regarding some of these limitations. For example, Foteinaki et al. (2019) propose a model to associate activities and appliances based on percentages of appliance ownership at a national level (related to L1). Yilmaz et al. (2017) introduce an empirical-based stochastic model of appliance use that benefit from monitoring spanning over several days (related to L2). Buttitta et al. (2019) proposed a multi-day occupancy model based on a Markov Chain process to generate weekly heating-load profiles (related to L5).

But to our knowledge there is no publication that presents a more consistent approach to all the TUS limitations. In line with (Schumann et al, 2021), who pointed out some issues related to classical occupant behavior models, Osman et Ouf (2021) highlighted the potential of integrating TUS with different sources of data that cover social, economic, and building aspects. They recommended to integrate TUS data with energy metered or surveyed data to fill any missing information and get a holistic view for the occupants' behavior and their impact on energy use profile. In addition, Berger and Mahdavi (2022) suggest that ABM can effectively capture the behavior of building occupants, both as individuals and groups. However, they note a deficiency in studies that examine the incorporation of ABM in building performance simulation. The authors underline the usual limitations associated with ABM: lack of detailed information about agents' behavior and their interactions, reproducibility issues, and difficulty to evaluate the fidelity of ABM results given their potential emergent complexity.

This paper proposes to answer these criticisms through a comprehensive approach linking qualitative and quantitative studies on human activities with ABM to address the simulation of energy consumption in the residential sector.

## Our combined approach of qualitative and quantitative studies and ABM

The focus of this paper is our approach to employing and supplementing TUS data (whereas a general presentation of our platform can be found in (Schumann et al, 2021). This work is rooted in a technological research program that considers conceptual, methodological, and technical aspects, including the relationship between activity models and multi-agent systems, qualitative and quantitative validation, and the development of a multi-agent simulator platform. First, we explain how qualitative empirical studies are essential for understanding human activity and tackling the aforementioned TUS limitations.

**The need for qualitative and situated approaches to human activity and ABM**

Qualitative studies of human activity favor conceptual and methodological frameworks that enable a detailed understanding of 'how' individuals and groups act in concrete everyday settings (e.g., Guibourdenche, 2013). Therefore, the SMACH project initially relied on the structural relationship between the qualitative studies of human activity and ABM to build its approach to activity modeling and its validation (Haradji et al., 2012). This first step notably meant articulating concepts, methods, and results from qualitative studies of activity in natural settings with incremental-situated ABM and social simulation. This work produced a basis for small-scale simulation (a few hours for one single household). This approach helped building coherent Agent Based (AB) sequences of individual actions in interaction with the other members of the family and the environment, at the household scale. It only then became apparent that TUS were necessary when we needed to work at a larger scale (thousands of households over several months of simulated time).

Qualitative studies allow domain experts to understand and explain the various forms, structures, dynamics, emergence, construction, or meaning of in situ activity. A wide range of approaches to human activity exists in the social sciences and ergonomics. These works require various qualitative methods, e.g., eliciting the action's meaning, observing practices, and analyzing conversations or verbalizations. The situated ABM approach is partly inspired by these frameworks, while being a reduction of activity analysis for ABM. It consists in placing a human participant in a simulated situation and observing their responses instead of collecting general information through an interview. The ABM is adjusted through a step-by-step approach based on the knowledge of the participant's real activity (Sempé et al., 2010).

Derived from these studies, three principles formed the qualitative basis for the SMACH project: 1) the individual agent are autonomous in realizing their tasks, 2) the collective activity is built in the dynamic interaction between the individuals (forming small and ephemeral sub-groups in the home), and 3) there is an asymmetric structural coupling between individuals/groups and their environment (devices, building, outdoor conditions).

This approach provides solutions to the TUS limitations outlined in the first section. Concerning L1 ("no link between activities and energy use"), the structural relationship between qualitative studies and ABM can help make accurate links between activities, devices, and energy use. (Guibourdenche, 2013) empirically described the various contexts of inhabitants' activity (individual, collective, and coupling to the material environment) in which the devices are used or not, forgotten or intentionally left on. He formally described the activity contexts of energy consumption by focusing on the parallel and interdependent concerns of individuals in action at home, e.g., a mother ironing and watching TV while coordinating the activity of the rest of the family in order not to be late for upcoming children's baths and the rest of the evening. This kind of study provides qualitative arguments when simulating the variations of activities, along with situated modeling and social simulation. It aligns with Shove et al. (2012)'s notion of practices as building time interdependently and with the approach considering "what energy is for" (Shove and Walker, 2014). Both standpoints of qualitative studies of action and situated ABM imply relating L2 (lack of variability and interactions) and L5 (TUS's biases): we need to use concepts and methods that enable us to describe the variety (richness) of action meaning and the individual-household-environment interactions.

Qualitative approaches consider temporal scales of activity ranging from a tenth of a second to several years. They are not limited to TUS's 10 minutes episode. Capturing short or more extended activities depends on the researchers' choice or the modeler's choice in situated modeling. As a result, they help characterize and explain activity's variations in real life. In a study focusing on second-to-day long activities, (Guibourdenche, 2013) shows how a mother moved the pile of laundry to be ironed the evening before this ironing (on a Wednesday) from the first floor of the house to the sofa in the living room on the ground floor, thus preparing for the ironing planned for the next day (Thursday). The next day (Thursday), in a period from 4:00 p.m. to 5:57 p.m., the processing of the clean laundry carried out by this same person is suspended 18 times. These suspensions are explained by the other actions and concerns to be realized by the mother, particularly caring for the children. These kinds of results helped frame explanations and produce coherent sequences for ABM, although the AB model is a substantial reduction of real-life activity.

L4 ("absence of collective activities or household-level organization") is another area where qualitative studies, situated modeling and social simulation have already described many phenomena concerning household energy consumption and social organization. As an illustration, a model exists for considering the precise degree of collective similarity and convergence between individual concerns in a household (Haradji et al., 2018). Situated modeling and social simulation help frame and adjust AB models to household dynamics in real life given technical possibilities and limits for reduction into a model. For example, Sempé et al. (2010) demonstrated how to create variations in the individual or collective realization of breakfast. L3 remains a more practical issue, depending on the researcher's choices about the population to be studied. Nevertheless, studying large categories from a situated and qualitative perspective on activity would call for further research.

Although further research is still necessary, several of the TUS limitations identified by McKenna et al. (2022) have been addressed in the SMACH project with the help of this structural relationship between ABM and the qualitative studies of real human activity: activities in parallel (individual or collective); short activities; consumptions as the indirect effect of collective and interdependent activities resulting from complicated household dynamics; coherent sequences of activities; explanation about sequences of activities and days. If "the concern is now not just who, but what is acting, and how" (Malik et al., 2022), qualitative and situated approaches to human activity and ABM can provide new opportunities for ABM and better comprehension of energy consumption beyond the limits of TUS. The main challenge remains to combine the concepts, methods, and results of qualitative and quantitative approaches to human activity and energy.

**A human activity model based on ABM**

We now introduce the agent model and how it benefits from multi-source qualitative and quantitative data. The model we present consists of two distinct modules: a population generation module and an activity module.

The population generation module, based on a classical model from the synthetic population research field (Müller and Axhausen, 2010), allows the generation of a synthetic population that exhibits the same characteristics as a targeted population. Based on large scale national studies (e.g. the "2014 French population census" and the "2017 housing survey" from the French National Institute of Statistics and Economic Studies), the module generates up to tenths of thousands of individuals (characterized by their gender, age, PCS, income, etc.), gathered in households (characterized by their family type, size, energy tariff, etc.), installed in dwellings (characterized by their surface, type, insulation, localization, weather), and equipped with appliances (heaters, water heater, light bulbs, home appliances, electric vehicle, solar panels, etc.). These features are inter-dependent and will impact the simulated load curves. For instance, the household's size is related to the dwelling's floor area, and both impact the hot water tank volume, whereas the localization affects the presence probability of an air conditioner. The population generation module opposes L3 directly.

The activity module notably addresses L2. In the best cases, TUS data contain 2 days for the same individual, but these days are non-consecutive and of a different type: a weekday and a weekend day. This makes it impossible to infer how much an individual acts similarly (or dissimilarly) from day to day. Yamaguchi & Shimoda (2017) use the notion of "routine" and "non-routine" behavior, which offers a promising approach to address L2. However, further data are required to support this model, specifically in terms of establishing criteria for distinguishing between routine and non-routine behaviors. State-of-the-art approaches have two classical options to deal with TUS data: either real timetables are directly copied and simulated (which leads to fixed scenario patterns), either these data are aggregated by "types of individual" (e.g. by sex, age, PCS, income, etc.) to represent their "average days". Both cases logically lead to a severe lack of variability of activity.

We propose to overcomes this impasse by adopting a third alternative: an ABM wherein simulated individuals are autonomous agents assigned with daily tasks that must be accomplished to the best of their abilities, depending on the context. These tasks are computed from TUS data and have several characteristics, such as a preferred period (the time frame during which the activity should preferably take place), duration range (minimum and maximum acceptable durations), frequency (number of repetitions during a day or a week). At each 1-minute timestep, simulated individuals replan their behavior by computing the priority of all their undone tasks and selecting the one with the highest priority. The priority computing takes numerous factors into account: the current task (agents prefer finish ongoing tasks rather than starting a new one); the other tasks they should be doing at the present time (i.e. all activities whose preferential period includes the current time step) because the more an agent is pressured by time, the less time it has for each tasks; the time remaining in the preferred period compared to the minimum duration of the task, etc. Exceptional events and the environment are considered in that process. For example, the impact of weather was computed for each type of individual based on TUS data, thanks to its "weather" column. Good weather increases the mean duration of leisure outside tasks and decreases the mean duration of inside leisure ones. Consequently, even if the list of tasks and their characteristics are identical, simulated individuals will not perform the same activities in the same sequence with the same duration on two consecutive days. This particularity ensures a first answer to L2: the activity variability comes from the ABM itself.

The task generation process, detailed in (Reynaud et al, 2017), ensures coherence by extracting task sequences from real timetables. Variability is achieved by using aggregated data from individual types to select task time characteristics such as duration and preferred period. Minimum and maximum durations as well as preferred periods for each activity are computed such that they represent X% of the data, centered around the mean. For example, if X=50%, it means that, for a specific type of individual, for a specific activity, for a specific type of day, 50% of all repetitions of this activity in the data have a duration between the minimum and the maximum duration, centered on the mean duration. Thus, it is possible to increase or decrease the activity variability by increasing or decreasing X. If X=100%, simulated individuals are free to choose the duration of each activity if this duration exists in the data. As X approaches 0, the simulated individuals become more constrained to perform tasks closer to their mean duration. As an example, for the "work" activity of active women over 50 during the week, if X is 90%, the preferred period is from 6:00 AM to 7:50 PM, with a duration between 2 and 12 hours (representing 90% of activities in TUS data). If X is 50%, the preferred period is shortened to 7:40 AM to 4:50 PM, with a duration between 5 and 10 hours.

Regarding L4, some previous models (e.g., Yamaguchi et Shimoda, 2017) worked toward the modelling of interactions among household members, but these interactions are often stereotyped (e.g., mandatory shared meals or no simultaneous use of the bathroom). In our model, we use the "who is present" column in TUS data to determine the activity's collectivity level: the percentage of times that this activity was carried out with other people. This allows for a priority bonus to be applied to other individuals when one is performing the activity, which increases activity variability. For instance, our simulations show that the "cooking" activity is performed with others about 60% of the time, while "housekeeping" is only done with others about 2% of the time.

Some decisions are made at the household level. Washing clothes should be considered collectively on a weekly basis, as the number of washing machine cycles required depends on household size rather than individual usage. Another example is the cooking activity. Cooking for a meal taken together should be considered at a collective level since everyone does not only cook for themselves; if someone cooks, potentially everybody can eat.

**Appliance use and energy consumption model**

We now introduce how the model and data were used to improve the appliance use model. Allocating appliances to activities and setting the probabilities of use (PU) is challenging because of the L1 and L5 TUS limitations. While obvious associations can be made, for instance, "watching TV" that requires a TV, TVs can also be turned on much more often than just during the "watching TV" activity (e.g., when used for background visual and sound). Some activity-appliance associations are more probabilistic in nature (e.g., people do not use the vacuum cleaner for all their housekeeping activities, nor do they use it for the whole duration of the activity). Moreover, some real activities do not appear in the TUS at all because their timespan is lower than the 10 min threshold or are not considered to be a main activity.

In SMACH, L1 is tackled by associating each task with one or several Appliance Use Models (AUM), that define the probability of a given appliance to be used during the task, and its operating mode. A task can therefore trigger the use of several appliances. To reflect the variety of real appliances and the way they are used, we introduce three types of AUM: Forced, Fractional and Cycle. In the Forced AUM, the appliance is used during the entirety of the task duration (e.g., "watching TV" leads to the TV being used for the entire duration of the viewing). In the Fractional AUM, the appliance works during part of the task and its use is scattered over the task duration (e.g., a vacuum cleaner whose use is disseminated during the housekeeping period). In the Cycle AUM, a cycle of the appliance is initiated, meaning that the appliance is used during a predetermined period that is not dependent on the duration of the task. This mode is applied to washing machines, dryers, dishwashers, and ovens. Some other appliances do not rely solely on human activity like heating or refrigerators and are not controlled by an AUM. Their thermal model have setpoint temperatures. Activity influences their operation, for instance through setpoint modifications for heating or refrigerator openings during meal preparation. The use of these AUM has the following advantages:

- Seasonality: PUs can be changed based on the time of the day, type of day, or the season. For instance, dryers are documented to be much more used during winter than summer. Another example is the strong difference between Saturdays and Sundays, visible in the read load curve data used for the validation and presented below;

- Variability: PUs enable different activities to have distinct load profiles, involving various appliances;

- Readability/explainability: PUs are easy to understand and manipulate, as illustrated in the Use Case section.

To calibrate the appliance model PUs, we leveraged the CONSER study (EDF, 2016), which is based on a questionnaire completed by 4000 households regarding their electrical appliances and their appliance use habits (approximate duration and time of use within five daily periods). The resulting data are cross-referenced with technical information and research results on appliances power to estimate each appliance's yearly unit energy consumption (Binet and Cayla, 2018). We use these data to select the appliances that require explicit depiction based on their unit powers, annual unit energy and ownership rate. We grouped appliances that do not require an explicit model into composite appliances that represent sets of appliances used for similar activities like cooking, hygiene, or digital practices. Their energy usage is included in a baseline that varies with time. For example, the cooking appliances consist of six components: an electric oven, an electric cooking plate, a microwave oven, a coffee machine, a kettle, and the kitchen baseline.

The calibration of the AUM was performed with help of the reference results of the CONSER study, which provides energy consumption targets in TWh for each category of appliances at the French population scale. The data used come from the updated and enriched 2019 version of this survey conducted in 2016, the results of which are presented in Figure 1.

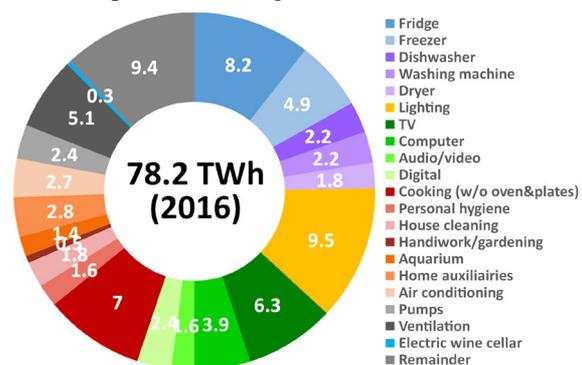

*Figure 1: Reconstruction of specific electricity consumption in TWh from the CONSER study*

These targets were studied in the model using a synthetic population of 1000 households statistically representative of the French population, their homes and appliances, and the current electricity tariff structure.

It appears that the targets in energy consumption cannot be satisfied if only the obvious association of activity and appliances are considered (e.g., "oven" only associated with "cooking"), even with a PU=1. It is therefore necessary to allocate AUM of some specific appliances to more activities than through a direct association. Examples of PUs for 3 appliances are shown in Table 1.

At the beginning of each task, the selection of the devices that will be used is done by a random draw based on their PUs. PU values are first estimated through expert considerations (the use of a coffee machine is more probable during breakfast than during dinner), and then refined with the help of real hourly power demand data.

*Table 1: PUs for a selection of tasks and appliances*

| Task | Microwave | TV | Computer |
|---|---|---|---|
| Cooking | 0.64 | 0.05 | 0.25 |
| Computer | 0.02 | 0 | 1 |
| TV | 0.02 | 1 | 0 |
| Reading | 0.02 | 0 | 0.06 |
| Housekeeping | 0.02 | 0.16 | 0.19 |
| Breakfast | 0.01 | 0.05 | 0 |
| Meal | 0.08 | 0.05 | 0.06 |
| Personal time | 0.01 | 0.16 | 0.19 |

Finally, an automatic calibration process is used to set the unit power of each appliance category based on the CONSER survey results. The parameters of the AUM are adapted to match the daily power reference data. For instance, in addition to the use of the CONSER results, the average cooking power at the population scale (29M households) was also adapted to fit the ADEME WattGo 2016 study (a study based on on-site measurements of appliance by appliance electricity consumption from 118 French households between 2014 and 2016):

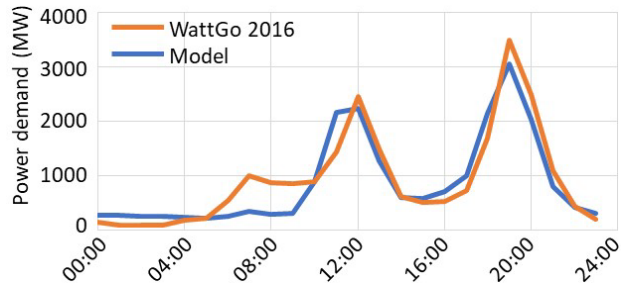

*Figure 2: Power of oven and cooking plate on an average day at the French population scale*

**Validation**

We now demonstrate how the model led to fitting results compared to real load curve data at an aggregated scale. The validation of our model on activity and energy consumption is a continuous work as described in (Schumann et al, 2021). A future publication will detail the validation including DHW and heating, while this paper focusses on specific electricity. The validation of aggregated power loads and energy consumptions was conducted by comparing model results with aggregated data from the "Panel Particuliers" (PP) panel of consumers. The data consist in French household power demand records at the timestep of 30 minutes, recorded for a maximum duration of 2 years. We used the answers of the participants about their home to generate a synthetic population representative of the panel (e.g., appliances, housing, inhabitants, habits such as holidays or weekend absences). We selected 300 households without electric heating nor electric DHW that exhibited power records of acceptable data quality and reliable questionnaires between March 2019 and February 2020 (pre-Covid). The 30min time-step averaged power demand for each month (i.e., the average of the 4 weeks of each month) were compared using a set of complementary metrics. The Mean Absolute Error (MAE) assessed the proximity between the model and the data; the Root Mean Square Error (RMSE) helped, along with the MAE, to identify the presence of extreme values. The Mean Absolute Percentage Error (MAPE) measured the percentage error of the forecast, while the Weighted Average Percentage Error (WAPE) was useful for identifying near-zero values. The Mean Directional Accuracy (MDA) was employed to compare forecast directions, and the Fréchet distance served as a measure of similarity. As an illustration the power demand for February 2020 is shown in Figure 3. The results, supported by the values of the RMSE in Table 2, demonstrate a proper dynamic of the aggregated power loads, including the minimums and maximums, as well as the ability to represent the distinct shapes of Wednesdays, Saturdays, and Sundays:

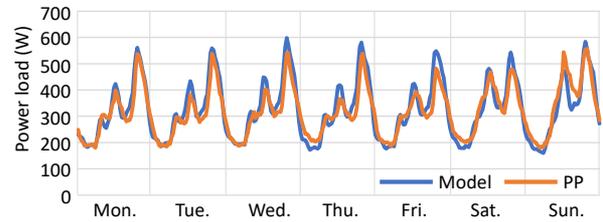

*Figure 3: Average power (mean week Feb. 2020)*

*Table 2: Monthly RMSE (in W) between Model and PP*

| Jan. | 37.2 | May | 45.6 | Sep. | 43.7 |
|---|---|---|---|---|---|
| Feb. | 39.2 | Jun. | 44.1 | Oct. | 38.7 |
| Mar. | 46.1 | Jul. | 32.5 | Nov. | 40.0 |
| Apr. | 47.6 | Aug. | 33.2 | Dec. | 43.4 |

We also validated the seasonality of power demand. Figure 4 shows the absolute values and relative deviation in average monthly power demand over the year:

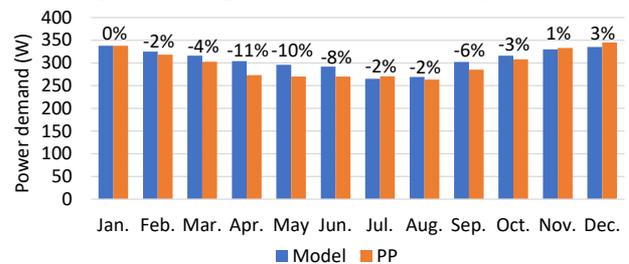

*Figure 4: Average monthly power demand (W)*

**Use cases and illustrations**

The model was used to quantify the impact of behavioral conservation measures for energy efficiency and peak shaving ("eco-behaviors") on the national load curve during the 2022-2023 winter energy crisis in France. We evaluated which behaviors (e.g., changes in showers, cooking, laundry) would be the most appropriate to reduce the power demand during the peak hours defined by the TSO (8:00-13:00 and 18:00-20:00).

The first Use Case is an eco-behavior consisting in refraining from using cooking appliances during peak hours. As stressed by Shove et al. (2012), practices constrain each other and shifting an activity will have an impact on the other activities. Families, particularly parents with young children, prefer not to postpone evening activities for too long and try to anticipate when they can (Guibourdenche, 2013). Those diverse behaviors can be achieved with our activity model by displacing the preferential periods (PP) of cooking tasks outside of peak hours, with a maximum shift of 45 minutes to represent the limited possible delay in evening activities. This leads to a diversity of behaviors among agents, some choosing to anticipate their cooking tasks, others opting to postpone them, and some maintaining their regular schedule, rather than enforcing a uniform behavior on all agents.

Figure 5 shows the effect of such behavior on the activity rate and the average cooking appliances power demand. It is drawn from a 1000 dwellings simulation. The 34 possible tasks in the model were grouped in 8 categories. Most of the cooking and eating tasks are delayed after 20:00 and few of them are done sooner. The released time is replaced mainly with leisure taks and idle (respectively an 8% and 9% rise at 19:00). "Idle" occurs when an agent does not have any specific task available. The power gain is substantial, reaching a maximum of 250W.

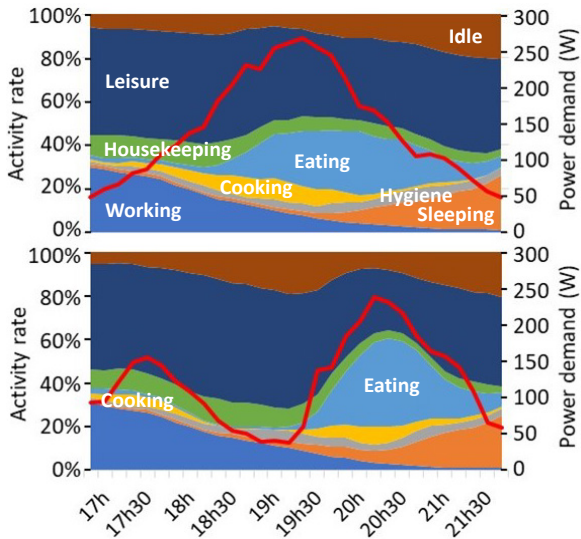

*Figure 5: Activity rate and cooking power load (red) without (top) and with (bottom) eco-behavior*

The second Use Case is a behavior aimed at avoiding taking showers and baths during peak hours to lower DHW power loads. In our model, showers and baths may be triggered during a hygiene task. Each simulated individual receives a fixed number of showers per week based on a reference survey (ADEME/COSTIC 2016). At the beginning of a hygiene task, the occurrence of a shower is randomly drawn considering the day of the week and the number of showers already taken during the current day and since the beginning of the week. The eco-behavior consists in setting to zero the probability of a shower during peak hours. As there are often several hygiene tasks during the day, agents are expected to take advantage of off-peak hygiene tasks to take their showers.

Figure 6 presents both the average hot water consumption and the average DHW power demand of a mean weekday with and without showers during peak hours. Hot water consumption is essentially transferred from peak hours to the evening. The reduction of power demand however is limited and occurs only in the morning, with a limited rebound effect at night. In France, a majority of electrical DHW are controlled to heat water only at night and in the beginning of the afternoon, when the power demand is low, which makes the relation between hygiene activities and power load very indirect.

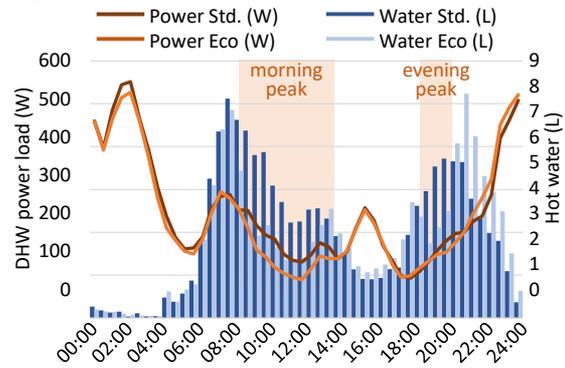

*Figure 6: DHW power (left) and hot water consumption (right) with and without the "No shower" eco-behavior*

This Use Case illustrates how the model is able consider the complexity of the link between activity and appliance use (L1) and to represent plausible behavioral modifications and the interactions between activities (L2).

## Conclusion and perspectives

To overcome the limits related to the TUS and their use (L1-L5), we presented an Agent-Based Model combined with qualitative and quantitative studies. L3 is addressed by the population generator. L2 and L4 are addressed via the agent-based activity model and the inputs of qualitative studies which also help understanding L5. L1 is addressed via the presented Appliance Use Models based on calibrated Probabilities of Use. The model proposes a holistic approach for the representation of human activity and improves the consistency of simulated activity and energy consumption of households. This model has immediate practical applications and was used to quantify the impact of behavioral flexibility measures.

Our work currently focuses on better understanding the cognitive and sociological macro-determinants of stability and variability of activity and how to integrate them in an ABM architecture. Future works will focus more on local scales such as neighborhoods or cities, by better considering the geographical, economic and socio-demographic factors of the studied territories. One other major challenge is the correct representation of the heating load curve at a national scale, considering heating practices and behaviors (including air renewal and air quality) such as those documented in the survey conducted in 2018 in France (Laurent et al 2022).